# Generative Adversarial Networks and Perceptual Losses for Video Super-Resolution

Alice Lucas, Santiago Lopez-Tapia, Rafael Molina and Aggelos K. Katsaggelos.

*Abstract*—Video super-resolution (VSR) has become one of the most critical problems in video processing. In the deep learning literature, recent works have shown the benefits of using adversarial-based and perceptual losses to improve the performance on various image restoration tasks; however, these have yet to be applied for video super-resolution. In this work, we propose a Generative Adversarial Network(GAN)-based formulation for VSR. We introduce a new generator network optimized for the VSR problem, named VSRResNet, along with a new discriminator architecture to properly guide the VSRResNet during the GAN training. We further enhance our VSR GAN formulation with two regularizers, a distance loss in feature-space and pixel-space, to obtain our final VSRResFeatGAN model. We show that pre-training our generator with the Mean-Squared-Error loss only quantitatively surpasses the current state-of-the-art VSR models. Finally, we employ the PercepDist metric ([2]) to compare state-of-the-art VSR models. We show that this metric more accurately evaluates the perceptual quality of SR solutions obtained from neural networks, compared with the commonly used PSNR/SSIM metrics. Finally, we show that our proposed model, the VSRResFeatGAN model, outperforms current state-of-the-art SR models, both quantitatively and qualitatively.

## I. INTRODUCTION

The task of video super-resolution, which corresponds to estimating high-resolution (HR) frames from their observed low-resolution (LR) versions, has become one of the central problems in image and video processing. With the growing popularity of high-definition display devices, such as High-definition television (HDTV), or even Ultra-high-definition television (UHDTV) on the market, there is an avid demand for transferring LR videos into HR videos so that they can be displayed on high resolution TV screens, void of artifacts and noise.

The objective posed by the Video Super-Resolution (VSR) problem is to reconstruct a high-resolution sequence $\{x_1, x_2, ..., x_{T-1}, x_T\}$ given a corresponding low-resolution sequence $\{y_1, y_2, ..., y_{T-1}, y_T\}$. Algorithms which tackle the SR problem can be divided into two broad categories: model-based and learning-based algorithms. In model-based approaches (e.g., [3], [4], [5], [6]) the low-resolution (LR) frames are explicitly modeled as blurred, subsampled, and noisy versions of the corresponding high-resolution (HR) frames, i.e. $y_i = DHx_i$ where $x_i$ is the $i$-th high-resolution frame in the sequence, H is the blurring operator, D the downsampling matrix, and $y_i$ the corresponding observed low-resolution frame. With this explicit modeling, one can invert the SR model to obtain an estimate of the reconstructed HR frame. Due to the strongly ill-posed nature of the SR problem, careful regularization must be used when solving for the reconstructed frame. Signal priors must be used to enforce image-specific features into the HR estimate. For example in the Bayesian framework, priors controlling the smoothness or the total variation of the reconstructed image are utilized to regularize the SR problem (see for example [3], [4], [5]).

On the other hand, conventional learning-based algorithms do not explicitly make use of the analytical SR model and instead use large training databases of HR and LR videos to learn to solve the video super-resolution problem. Recently, Deep Neural Networks (DNNs) have been proposed as another learning-based tool used for video super-resolution. In the general case of using deep neural networks for video SR, the goal is to find a function f(·) such that $x_t = f(Y_t)$. In other words, f(·) learns the mapping from the LR center frame and the corresponding past and future frames, e.g., $Y_t = (y_{t-k}, ..., y_{t-1}, y_t, y_{t+1}, ..., y_{t+k}), k \geq 0$, to obtain an estimate of the reconstructed center HR frame $x_t$.

The traditional approach to train DNNs for video super-resolution is to first artificially synthesize a dataset with corresponding high-resolution and low-resolution frames. The Mean-Squared-Error (MSE) cost function between the estimated high-resolution frame $x_t$ and the ground truth frame is then used as the cost function during the training of the neural network. Numerous works in the literature (e.g., [7]) have shown that while the MSE-based approach provides reasonable SR solutions, its fairly conservative nature does not fully exploit the potential of deep neural networks and instead produces blurry images. As an alternative to the MSE cost function, literature for NN-based super-resolution has proposed the use of feature spaces learned by pre-trained discriminative networks to compute the $l_2$ distance between an estimated and ground truth HR frame during training. Using such feature-based losses in addition to the MSE loss has been proven to be effective at significantly boosting the quality of the super-resolved images.

Generative Adversarial Networks (GANs) [8] are powerful models which have been shown to be able to learn complex

Preliminary results of this work were presented at the 2018 IEEE International Conference on Image Processing (ICIP) [1]. This work was supported in part by the Sony 2016 Research Award Program Research Project. The work of SLT and RM was supported by the the Spanish Ministry of Economy and Competitiveness through project DPI2016-77869-C2-2-R and the Visiting Scholar program at the University of Granada. SLT received financial support through the Spanish FPU program. A. Lucas and A.K. Katsaggelos are with the Dept. of Electrical Engineering and Computer Science, Northwestern University, Evanston, IL, USA. S. Lopez-Tapia and R. Molina are with the Computer Science and Artificial Intelligence Department, Universidad de Granada, Spain. **This work has been submitted to the IEEE for possible publication. Copyright may be transferred without notice, after which this version may no longer be accessible.**

distributions by sampling from these with the use of deep neural network. Originally introduced in the context of image generation ([8]), GANs have since been used for a multitude of generative tasks, such as various image-to-image translation tasks, 3D modeling, and audio synthesis. The generative ability of these models has been exploited to produce images of exceptionally high quality for several image reconstruction tasks (e.g., [9], [10], [11]). While GANs have been applied to the image super-resolution in numerous ways (e.g., [10]), they have not been applied to the problem of video super-resolution yet. Similarly, the use of feature-based losses for video super-resolution still lacks in today's literature. Therefore in this paper, we extend the use of GANs and feature-based loss functions to the intricate problem of video super-resolutions with deep neural networks.

The rest of the paper is organized as follows. We provide a brief review of the current literature for learning-based VSR in Section II. In Section III-A, we introduce a residual architecture for video super-resolution, denoted VSRResNet, which surpasses current state-of-the-art algorithms and removes the need to apply motion-compensation on the input video sequence. Next, in Section III-B we re-frame the VSRResNet architecture in an adversarial setting. In addition to using an adversarial loss, we add feature-based losses in the overall cost function. The training procedure and experiments which provide the resulting VSRResFeatGAN is explained in more detail in Section IV. In our final section, Section V, we evaluate the performance of VSRResFeatGAN by comparing it with the current state-of-the-art learning-based approaches for video super-resolution for scale factors of 2, 3 and 4. Using quantitative and qualitative results, we show that our proposed VSRResFeatGAN model successfully sharpens the frames to a much greater extent than current state-of-the-art deep neural networks for video super-resolution.

## II. Related Work

In the past couple of years, multiple DNN-based models for video SR have been proposed in the literature. Liao et al. [12]'s approach follows a two-step procedure in which an ensemble of SR solutions is first obtained through the use of an analytical approach, and then used as input to a Convolutional Neural Network (CNN). Kappeler et al. [13] design an end-to-end approach and instead learn a direct mapping between the bicubically interpolated low-resolution frames, $Y_t$ and the corresponding central high-resolution frame $x_t$. Other works have experimented with the use of Recurrent Neural Networks (RNNs) for video super-resolution, for example in [14], where the authors use a bidirectional RNN to learn from past and future frames in the input low-resolution sequence. While RNNs have the advantage of explicitly learning the temporal dependencies in the input frame sequences, the challenges and difficulties associated with their training has led to CNN being the favored neural network for video super-resolution. In this direction, Li and Wang [15] show the benefits of residual learning with CNNs in video super-resolution by predicting only the residuals between the high-frequency and low-frequency frame. Caballero et al. [16] jointly train a spatial transformer network and a CNN to warp the videos frames to one another and benefit from sub-pixel information. Similarly, Makansi et al. [17] and Tao et al. [18] found that performing a joint upsampling and motion compensation (MC) operation increases the SR performance of the model.

Each of these models use the MSE loss as the guiding cost function for training their neural networks, hence resulting in estimated HR frames which are still fairly blurry. In the field of image super-resolution, the use of feature-based losses as additional cost functions, along with the use of GAN-based frameworks for training has been shown to result in significantly superior HR estimates compared with the ones obatined with traditional NN-based frameworks, such as the ones described above. For example, Johnson et al. [7] found that the use of feature-based loss as a loss function for learning the super-resolution task significantly increases the sharpness of the estimated HR image. Ledig et al. [10] were the firsts to use a GAN network and feature losses for learning to super-resolve images, which produced images with a previously unseen photorealistic quality.

## III. Adversarially Trained Deep Residual Architecture for Video SR: VSRResFeatGAN

In this section, we first describe a novel neural network architecture, VSRResNet, to solve the task of video super-resolution. Next, we re-frame the VSRResNet architecture in a GAN-based setting to further increase the perceptual quality of the super-resolved frames. Finally, we describe the use of feature-space and pixel-space loss functions to further improve the performance of our VSR model.

### A. The VSRResNet architecture

While single image super-resolution algorithms have used very deep neural networks to improve their model, this approach has not been applied yet to VSR. We argue that adding depth to the model increases the capacity of the model, which in turn provides better learned solutions for the VSR problem. To increase the depth of a model and avoid the vanishing gradient problem, we choose to design an architecture based on a chain of residual blocks, resulting in a neural network composed of a total of 34 convolution operations. The details of the architecture are shown in Figure 1. Our proposed architecture, VSRResNet, is based on a series of residual blocks, each composed of two convolution layers with learnable kernel of size 3 × 3. A Rectified Linear Unit (ReLU) activation function follows each convolution step. As shown in Figure 1, the VSRResNet architecture is explicitly designed in order to extract spatial information from each input frame and then fuse the information together. More specifically, the first convolution layer applies a convolution operation individually to each of the five frames in the input sequence. We performed an experiment in which we instead stack the input frames together (early fusion) and then apply a convolution operation to these concatenated frames. In this case we observed a small decrease in the PSNR performance of our network and therefore we did not adopt such an early fusion approach (see also [13] for more experiments on early and late fusion of architectures for VSR). The second convolution operation takes a concatenation of the extracted features across the different time steps to fuse the information from the previous step. The following fifteen residual blocks then learn the transformation that provides the final HR solution. We note here that we also experimented with smaller and larger numbers of residual blocks to determine our final VSRResNet architecture. More specifically, we found out that by using 5, 10, or 20 residual blocks instead of the proposed 15 residual blocks, the PSNR on our test dataset decreases by 0.90 dB, 0.20 dB, and 0.36 dB, respectively. Similarly, to determine the best number of input frames to the VSRResNet, we modified its architecture to acccept as input either 3 or 7 frames, instead of the proposed 5 input frames. We found out that these architectural changes resulted in a PSNR decrease of 0.19 dB and 0.72 dB, respectively, which suggests that using 5 input frames provide the optimal performance for our task.

As we later show in Section V the increased depth in VS-RResNet provides the network with more capacity to learn from the motion in the input frames and produce higher quality frames. Therefore, unlike most state-of-the-art systems for video super-resolution which perform motion-compensation on the input video, we choose to train the VSRResNet architecture on a non motion-compensated dataset, to let the network extract useful information from the motion. In addition to learning from motion, not using motion compensation provides the additional benefit of significantly reducing the computational time of the proposed method.

In the next section, we include the VSRResNet architecture as part of an adversarial framework with perceptual losses. We call our resulting model the VSRResFeatGAN model.

### B. The proposed adversarial system

Generative Adversarial Networks (GANs) [8] learn to generate samples from a specific data distribution through an adversarial training procedure. In the traditional GAN approach for image generation, a *generator* network learns to generate an image given a latent random vector z at its input. The learning of the generator is guided by an auxiliary network, a *discriminator*, which is simultaneously trained to distinguish between the images generated by the generator from images from the training dataset. Given a generator $G(z)$, on latent variables z to be later defined, the discriminator is trained to distinguish between real and fake images, i.e. output $D(x) = 1$ when x is sampled from the training dataset of natural images and $D(G(z)) = 0$ when the images are produced by the generator. On the other hand, the generator is trained to make the discriminator believe that its generated images $G(z)$ are real, i.e., trained to assign the discriminator output a probability $D(G(z)) = 1$. As a result of this adversarial training, the generator eventually converges to a solution which the discriminator fails to identify as "fake", which generally implies successful learning of the image manifold by the generator.

Adapting the original GAN framework to the problem of video super-resolution, we propose to use the powerful generative property of GANs by training a GAN to super-resolve high-resolution center patches from a given input sequence of low-resolution frames. Using a GAN-based training instead of the MSE-based training enables the model to obtain frames of much higher perceptual quality. We modify the original GAN setting by inputting the sequence of input low-resolution frames Y to the generator instead of a random vector z. This is similar to the use of GANs in still image super-resolution ([10]), in which case a single low-resolution image is provided at the input of the generator. The generator is adversarially trained to super-resolve the input LR frames such that the discriminator cannot distinguish between the reconstructed HR frames, $\hat{x} = G(Y)$ and those obtained from the training dataset. To this end, we use the GAN formulation first introduced in [8] and adapt it to video super-resolution by solving:

$$\min_{\theta}\max_{\phi} \mathcal{L}_{\text{GAN}}(\phi,\theta) = \mathbb{E}_x[\log D_\phi(x)] + \mathbb{E}_Y[\log(1 - D_\phi(G_\theta(Y)))] \quad (1)$$

where x is the center high-resolution frame of dimensions $N \times N$, Y is the sequence of low-resolution input frames around its low-resolution version y, each of dimensions $N \times N$, $D_\phi$ is our discriminator with trainable parameters $\phi$ and $G_\theta$ is the generator network with trainable parameters $\theta$, where here these parameters correspond to the learneable convolutional kernels of our networks.

We fix the architecture of the generator network to the VSRResNet architecture in Figure 1. The proposed discriminator $D_\phi$'s architecture is shown in Figure 2. It is composed of three convolution layers followed by a fully connected layer and sigmoid operation, which provides the probability of a real patch. We also experimented with the use of a very deep CNN as the one defined in [10], but found that the large capacity of this latter discriminator prevented the subsequent learning of our generator. Thus, for all the experiments in this paper, we fix the discriminator architecture to be the one in Figure 2.

### C. Adding feature-space and pixel-space distance as regularizers

Using the GAN function in Eq. 1 alone usually results in strong artifacts in the estimated super-resolution high-resolution frame, $G_\theta(Y)$. Examples of what the resulting super-resolved frames may look like are shown in Figure 3. From this figure, it is clear that the use of the adversarial loss alone results in the generator network learning to produce high-frequency artifacts, which resemble ringing patterns around the edges of the frame. In order to regularize such undesired effect of the adversarial loss, it is necessary to add regularizers to the adversarial Equation 1. In today's literature, commonly used regularizers for the GAN loss correspond to distances between the estimate $\hat{x} = G_\theta(Y)$ and the ground truth x available from the training dataset. The first regularization term measures the distance between the estimated $\hat{x}$ and the ground truth x in pixel-space, whereas the second term provides the distance in a pre-defined feature space. Here, the "distance" is provided by the Charbonnier loss, defined as:

$$\gamma(\hat{x}, x) = \sum_i \sum_j \sqrt{(\hat{x}_{i,j} - x_{i,j})^2 + \epsilon^2} \qquad (2)$$

where $i, j$ denote the pixel coordinates and $\epsilon$ is a small constant close to zero, which for our experiments we set to $\epsilon = 0.001$. The Charbonnier loss may be seen as an approximation to the $l_1$ loss. We found that using the Charbonnier distance instead of the traditional $l_2$ loss in pixel and feature-space as regularizers for the GAN training improves the learning behavior of the GAN model. We explain our findings in more detail in the experiments section, Section IV. The Charbonnier loss in pixel-space provides regularization in pixel-space, to ensure that the super-resolved frames does not depart by a great extent from the content in the corresponding ground truth high-resolution frame. The second term, the Charbonnier regularization in feature-space, leverages the deep features learned by deep discriminative classifiers to compare the the reconstructed frame from the ground truth frame. We choose our feature space to be the representation space obtained from extracting the feature maps from the third and fourth convolution layer of the VGG network defined in [21], denoted as VGG() in this paper. We found from our experiments that these two loss components are necessary for producing HR frames of high perceptual quality. Thus our proposed framework for training our VSR system becomes $\min_\theta \max_\phi L_{total}(\phi, \theta)$ where:

$$
\begin{aligned}
L_{total}(\phi, \theta) = \; & \alpha \sum_{(x,Y) \in T} \gamma(VGG(x), VGG(G_\theta(Y))) \\
& + \beta[\mathbb{E}_x[\log D_\phi(x)] + \mathbb{E}_Y[\log(1 - D_\phi(G_\theta(Y)))]] \\
& + (1 - \alpha - \beta) \sum_{(x,Y) \in T} \gamma(x, G_\theta(Y)) \qquad (3)
\end{aligned}
$$

where x and Y are sampled from the training dataset $T$, VGG(x) and VGG($G_\theta(Y)$) denote the feature maps obtained by providing x and Y as the input to the $VGG$ network, and the weights $\alpha > 0$ and $\beta > 0$ with $\alpha + \beta < 1$ are hyper-parameters which control the contribution of each loss component and are determined experimentally.

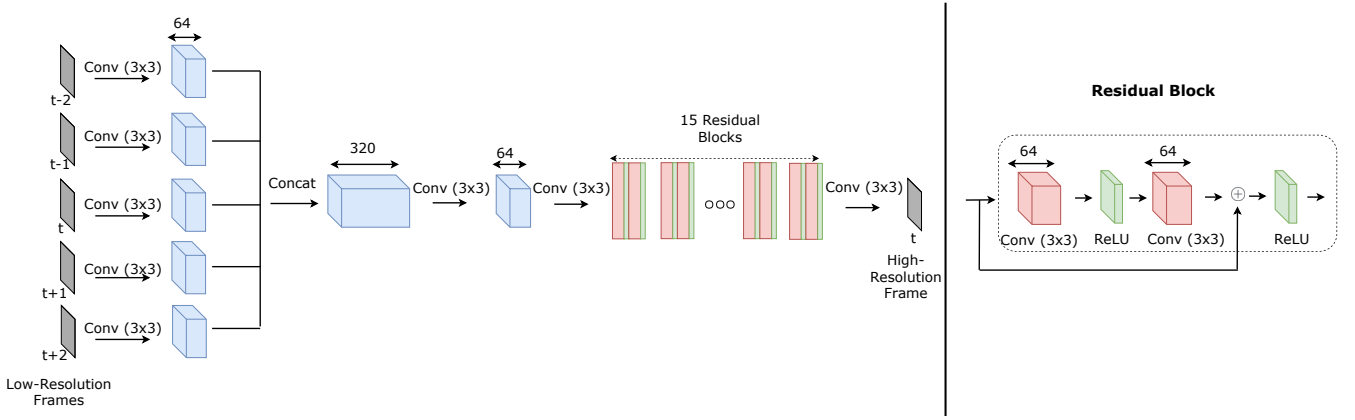

**Figure 1:** The proposed VSRResNet architecture. The network consists of a series of convolution operations with 64 kernels of size $3 \times 3$, applied on each input frame. The resulting feature maps are then concatenated together to obtain 320 feature maps. This is followed by two convolution operations and 15 residual blocks. Each residual block consists of two convolutional operations with 64 kernels of size $3 \times 3$, each followed by a ReLU layer. Following the definition of a residual block, the inputted feature maps are added to the output feature maps to obtain the final output of the residual block.

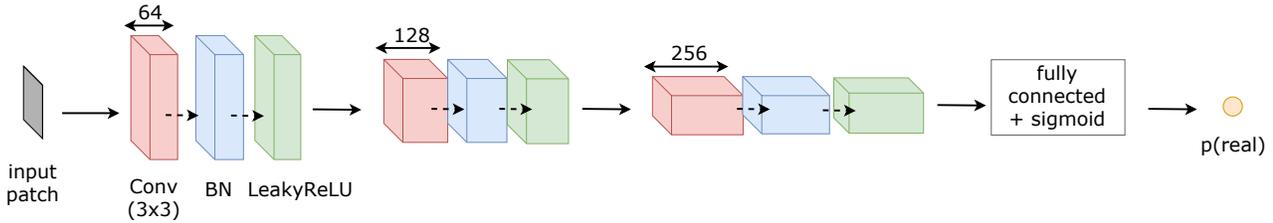

**Figure 2:** The proposed discriminator architecture architecture. The input to the discriminator corresponds to either an HR patch x from the ground-truth dataset, or a patch provided by the generator $G_\theta(Y)$. Its output corresponds to the probability that the input is a real HR patch. All convolutions in the discriminator use $3 \times 3$ convolutional kernels. The number of convolution feature maps used for each convolution step corresponds to the number shown on top of each convolution data cube (e.g., 64, 128, or 256). Each convolution layer is followed by a Batch Normalization (BN) layer as defined in [19] and the LeakyReLU operation defined in [20].

We name the resulting model the VSRResFeatGAN model. In the next section, we provide the details of our training procedure used for training VSRResFeatGAN.

## IV. Experiments

In this section, we describe the steps taken towards the training of our final model, the VSRResFeatGAN model.

### A. Training Dataset

To synthesize our training dataset of HR/LR pairs, we use the Myanmar video sequence, which was obtained from a publicly available video database ([22]). The Myanmar video contains 59 video sequences, of which 53 were used for training, and 6 for testing, following Kappeler et al.'s [13]'s approach. While the Myanmar videos in their raw and uncompressed form are 4K resolution ($3840 \times 2160$ pixels), we downsampled the frames in each scene by four to obtain frames of resolution $960 \times 540$. For each frame at time $t$, the high-resolution patches were obtained by extracting $36 \times 36$ patches from the HR frame. To synthesize the corresponding LR patches, we first performed bicubic downsampling on the HR frames at times $t-2$, $t-1$, $t$, $t+1$ and $t+2$, followed by bicubic interpolation on these frames. We then extracted the corresponding LR patches. at times $t-2$, $t-1$, $t$, $t+1$ and $t+2$. All of the downsampling and interpolation operations use MATLAB's *imresize* function. As a result, our training dataset consists of near 1 million HR/LR pairs, where for each ground truth $36 \times 36$ HR patch $x_t$ in our dataset, we are provided with a sequence of the corresponding five corresponding $36 \times 36$ low-resolution patches $Y_t = \{y_{t-2}, y_{t-1}, y_t, y_{t+1}, y_{t+2}\}$.

### B. Pre-training of the generator architecture

Clearly, the generator's task of learning a super-resolution function $G$ that accurately super-resolves LR patches is a much more difficult task than that of the discriminator, which is given the objective to discriminate reconstructed patches from ground-truth high-resolution patches. If the generator produces patches with significant artifacts, the discriminator's task then becomes trivial and failure of GAN training may follow. Therefore, it is critical to have the generator network start at a reasonable point in the beginning of the training, to ensure proper convergence of the generator and discriminator's loss functions. To this end, prior to starting the adversarial training, we first train the VSRResNet neural network with the traditional MSE loss function:

$$L_{RMS}(x, G_\theta(Y)) = \|x - G_\theta(Y)\|_2^2 \qquad (4)$$

We train the VSRResNet model with the loss function in Equation 4 for 100 epochs using the ADAM [23] optimizer and a batch size of 64. The initial learning rate is set to 0.001 and is then divided by a factor of 10 at the 50th and 75th epoch of the training. We train the VSRResNet model for each of the SR scale factors of 2, 3 and 4. The training hyper-parameters are fixed across scale factors.

### C. Training of VSRResFeatGAN

We trained the VSRResFeatGAN model with a learning rate of $10^{-4}$ for both the generator and the discriminator networks. The weight decay was set to 0.001 for the discriminator and 0.0001 for the generator. Similary to the training of VSRResNet, batches of 64 patches were used to perform each gradient update with the ADAM [23] optimizer. We trained the VSRResFeatGAN for 30 epochs, which we found was a suitable number of epochs for achieving convergence of the GAN training.

## V. Results

In Section IV, we described our approach for performing DNN-based video super-resolution with perceptual losses in an adversarial training. In this section, we qualitatively and quantitatively assess the performance of our VSRResNet and VSRResFeatGAN models and compare these to the current state-of-the-art DNNs for video super-resolution. To quantitatively assess the perceptual quality of the frames estimated by VSRResFeatGAN, we propose the use of an additional metric for evaluating the perceptual quality of our super-resolving models, denoted as the PercepDist metric, which we describe in more detail in Section V-C.

### A. Evaluation of the effect of depth in VSRResNet

We first evaluate the performance of our proposed VSRResNet model, which corresponds to our deep residual architecture train with the Mean-Squared-Error loss only. To assess the effect of depth in our network, we compare its performance with its shallower counterpart, the VSRNet architecture, first introduced in [13]. The VSRNet architecture is similar to the VSRResNet architecture as it first extracts spatial information from the sequence of five input frames and then fused together with the subsequent convolution layers in the network. However, the number of convolution layer in VSRNet is limited to four, whereas our VSRResNet network contains 15 residual blocks, for a total of 34 convolution operations. Both networks were trained on the same training dataset. We report the computed PSNR and SSIM values obtained when evaluating the VSRNet and VSRResNet models on our Myanmar test frames in table I. Similarly to the generation of the training dataset, the Myanmar test frames were both downsampled using bicubic interpolation as implemented in MATLAB's *imresize* function and subsequently upsampled to their original $960 \times 540$ spatial extent to provide an initial estimate of the HR solution. The VSRNet results in the table are those reported by [13]. Note that while the VSRNet model was trained on motion-compensated dataset, the VSRResNet model was not.

It is clear from Table I that VSRResNet outperforms VSRNet by a large margin across all scale factors. This implies that a large boost in performance can be obtained by increasing the depth of the network with residual blocks. For qualitative comparison, we show in Figure 4 selected regions from the Myanmar test frames super-resolved by VSRNet and VSRResNet. By zooming in these regions, large

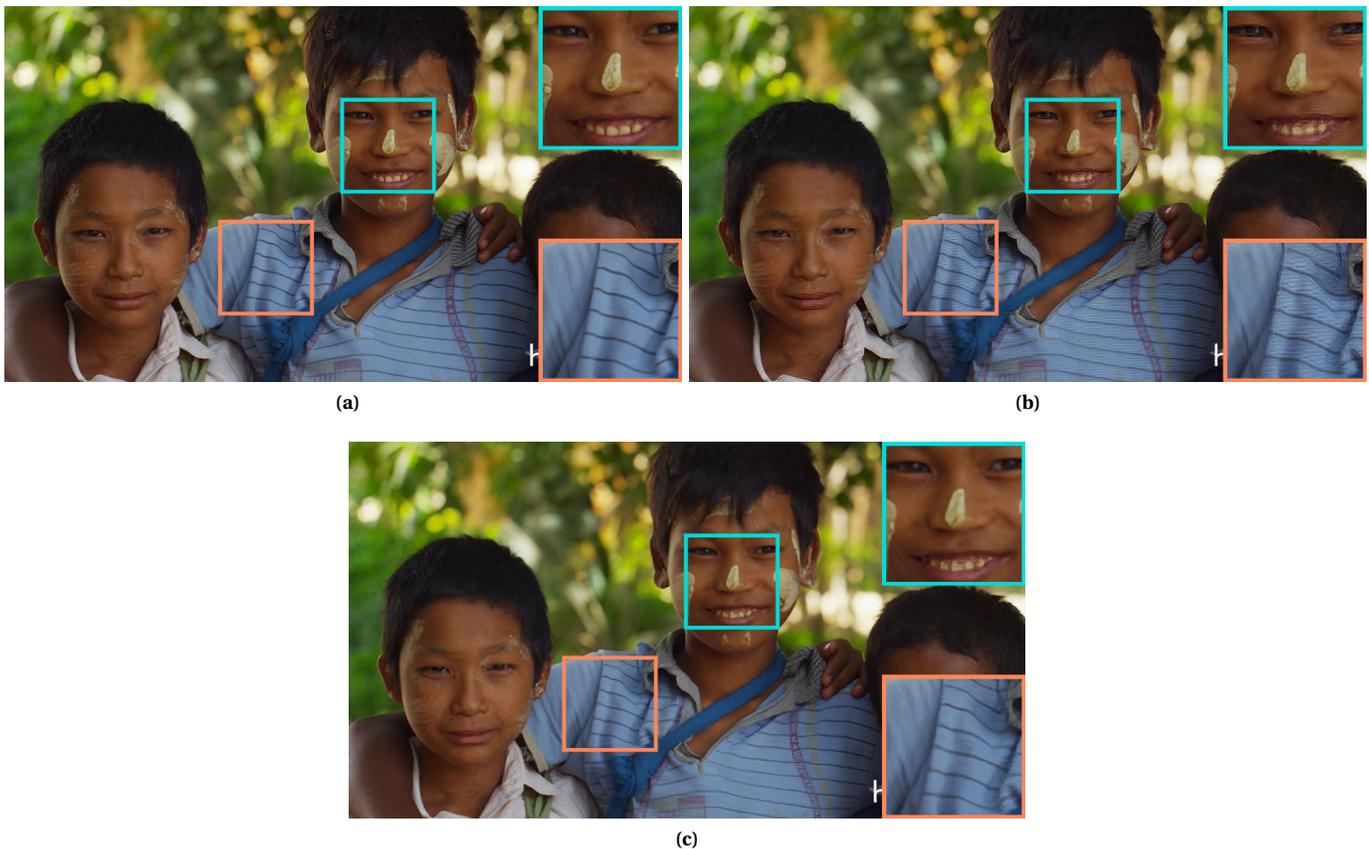

**(a)**  **(b)**

**(c)**

**Figure 3:** An illustration of the artifacts originating from using the adversarial loss alone for scale factor 3. (a) ground truth; (b) adversarial loss; (c) our proposed regularized adversarial loss.

| Scale | VSRNet (MC) | VSRResNet (non MC) |
|-------|-------------|--------------------|
|       | (PSNR/SSIM) | (PSNR/SSIM) |
| 2 | 38.48/0.9679 | 40.58/0.9807 |
| 3 | 34.42/0.9247 | 35.95/0.9481 |
| 4 | 31.85/0.8834 | 32.85/0.9075 |

TABLE I: Comparison of VSRResNet and VSRNet in terms of PSNR and SSIM. The evaluation metrics were computed on the Myanmar test dataset.

differences can be observed. As expected, the solution obtained by VSRResNet is of much sharper quality than the one provided by VSRNet, once again proving the benefits of increasing depth in a network trained for video super-resolution without having to motion compensate the input images.

### B. Learning from motion with VSRResNet

Having shown that adding depth provides the network with more capacity to learn accurate SR solutions, we now analyze whether the VSRResNet network successfully learns from the motion present in the sequence of input frames. Unlike most current VSR DNN-based algorithms, we chose not to perform explicit motion-compensation on the input LR sequence. This saves a considerable amount of processing time in real application and forces the network to learn to extract useful temporal information from the past and future frames and further improve the estimate of the center HR frame. To investigate whether the VSRResNet truly makes use of the motion present in the input, we design an experiment in which we replicate the center frame of the sequence across the five time steps, i.e. setting $Y_t = \{y_t, y_t, y_t, y_t, y_t\}$ and feed this "center only" sequence to VSRResNet. We compute the resulting PSNR values obtained by VSRResNet when evaluated on the Myanmar test frames and the VidSet4 test dataset [24] and report these in table II.

| Scale | EDSR [25] | VSRResNet | VSRResNet-cfo | difference |
|-------|-----------|-----------|---------------|------------|
|       | PSNR (dB) | PSNR (dB) | PSNR (dB) | $\Delta$PSNR (dB) |
| Myanmar-x2 | 39.09 | 40.58 | 37.54 | 3.04 |
| Myanmar-x3 | 35.34 | 35.95 | 33.86 | 2.09 |
| Myanmar-x4 | 33.13 | 32.85 | 31.45 | 1.40 |
| VidSet4-x2 | 29.93 | 31.87 | 28.92 | 2.95 |
| VidSet4-x3 | 26.37 | 27.80 | 25.67 | 2.13 |
| VidSet4-x4 | 24.68 | 25.51 | 23.85 | 1.66 |

TABLE II: Comparison of VSRResNet when using as input a sequence of frames versus the *center frame only* (VSRResNet-cfo). We report results for both the Myanmar and VidSet 4 datasets. The difference column is computed by subtracting the results of VSRResNet when using the center frame only from the results of using the whole input sequence.

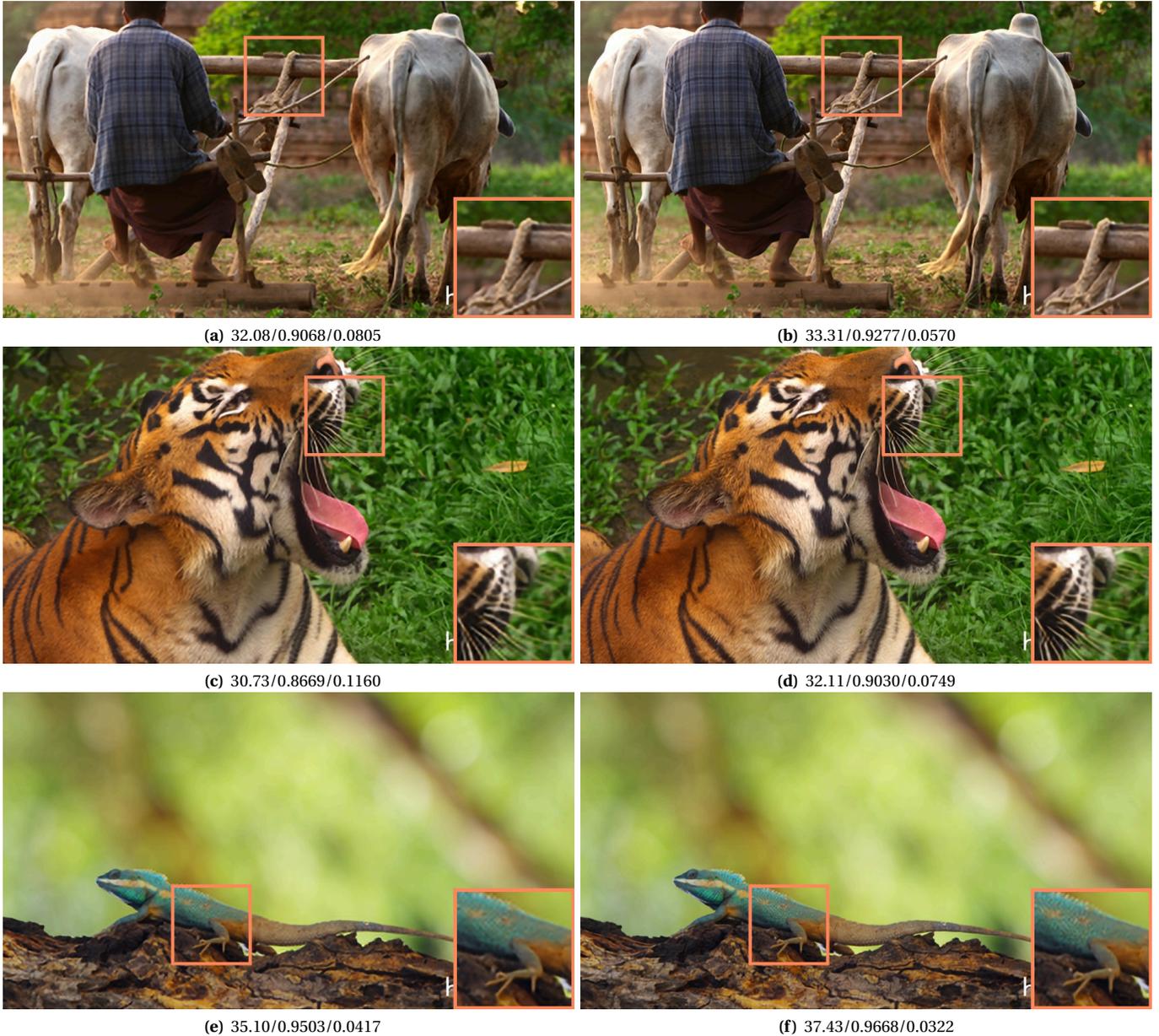

**(a)** 32.08/0.9068/0.0805

**(b)** 33.31/0.9277/0.0570

**(c)** 30.73/0.8669/0.1160

**(d)** 32.11/0.9030/0.0749

**(e)** 35.10/0.9503/0.0417

**(f)** 37.43/0.9668/0.0322

**Figure 4:** Qualitative comparison of VSRNet (figures (a), (c), and (e)) and VSRResNet (figures (b), (d), and (f)) for scale factor 3. The weights for the VSRNet model that produces these images were provided by the authors of [13]. By zooming in the selected regions large differences can be observed. In this figure and the rest of the figures in this paper, the three numbers in the caption below each frame correspond to the PSNR/SSIM/PercepDist metrics computed for that frame.

As shown in Table II, a large drop in the PSNR value is observed for both test sequences at all scale factors, more than 3 dB for the upscale factor of 2 on the Myanmar test dataset. This result suggests that VSRResNet successfully uses the motion in the input LR frame sequence to predict the center frame. We emphasize here that even though VSRResNet was trained on the Myanmar video dataset, which contains relatively small motion in between frames, it still efficiently uses the fast motion field present in the VidSet4 frames, as Table II clearly indicates. In this table we also present the results obtained from the EDSR network [25], which uses a deep residual architecture similar to ours, trained for super-resolving still images. We generated the super-resolved frames by EDSR [25] using the official Pytorch repository available at https://github.com/thstkdgus35/EDSR-PyTorch and computed the corresponding PSNR values in MATLAB. The results in Table II indicate that the EDSR network outperforms the VSRResNet for still image inputs quantitatively. However, for video sequences with motion it performs worse than the VSRResNet model that takes in *temporally consecutive* frames as input (except for Myanmar-x4). This experiment suggests two paths to improve our model: either to detect motionless sequences for which to use still image SR methods or alter its architecture to better handle motionless sequences.

*C. Comparison with state-of-the-art DNNs*

While the PSNR metric is the de facto standard metric for assessing the performance of an image restoration model, recent super-resolution literature has shown that the PSNR metric does not always provide an accurate assessment of the perceptual quality of images produced by deep neural networks. On the other hand, the family of discriminative Convolutional Neural Networks has been shown to learn features that seem to be positively correlated to the human's assessment of perceptual quality, [7]. This suggests that the representations learned by CNNs may be capable of providing us with a suitable metric for assessing the performance of an image restoration model. Recently, Zhang et al. [2] have trained a convolutional neural network to predict the perceptual similarity between a reference image and a distorted one [26], where the correct labels were obtained from large subjective studies. As a result of the training, the CNN learns to output a distance value between the ground truth and the distorted images. The smaller the distance between the two images, the better the perceptual quality of the distorted image. The authors found that the predictions provided by these neural networks agreed with the human judgement regarding the quality of a given image.

To accurately assess the quality of the images produced by our VSRResFeatGAN, we choose to use this metric, which we denote in this paper as the PercepDist metric. To compute the PercepDist metric, we use the github repository provided by the authors of [2], which contains the required pre-trained CNN model weights and evaluation code that computes the "distance" between two images. When using this metric for evaluating the performance of VSR systems, the image pair corresponds to the ground truth frame and the super-resolved frame outputted by our proposed model.

Equipped with this new metric, we now compare the performance of our VSRResFeatGAN model with multiple state-of-the-art VSR models. These include the VSR neural network proposed by Tao et al. [18], which is based on a convolution-LSTM neural network with efficient motion compensation on the input learned jointly within the network. In the rest of this paper, we refer to their work as SPMC-VSR. We also compare our model with two additional competing state-of-the-art VSR models: the VESPCN network proposed by [27] and the Temporal Adaptive Net proposed by [28]. The VESPCN network [27] incorporates temporal information into the VSR network by performing motion compensation on the past and future frames, and uses a sub-pixel convolution to pre-process the input frames in low-resolution space. The Temporal Adaptive Net [28] consists of a network with multiple SR branches, each responsible for super-resolving the frames at a temporal scale. A temporal modulation branch is then responsible for fusing the multiple VSR solutions into a single one. We also compare our model against two powerful SR models for still images: (1) the state-of-the-art image SR model proposed by Kim et al. [29], referred as VDSR in this paper, and the (2) SRGAN model [10], which uses a GAN-based loss similarly to our model. The VDSR network [29] corresponds to a very deep convolutional neural network with twenty layers trained to predict *residuals* between the low-resolution image and the unknown high-resolution image. The SRGAN model [10] is based on a deep residual neural network which was trained in an adversarial setting.

We test all models on the VidSet4 dataset, which was downsampled by factors of 2, 3 and 4. To test the SPMC-VSR model proposed in [18], we use the model weights and test code made available by the authors at https://github.com/jiangsutx/SPMC_VideoSR to super-resolve the VidSet4 sequence for scale factors of 2 and 4. The model weights for scale factor 3 being unavailable, we use the results reported in the paper [18]. The VidSet4 frames super-resolved by VESPCN [27] and the Temporal Adaptive Net [28] were found at https://twitter.app.box.com/v/vespcn-vid4 and http://www.ifp.illinois.edu/ dingliu2/videoSR/, respectively. To test the VDSR model proposed in [29], we use the code available at https://github.com/twtygqyy/pytorch-vdsr, and use their provided model weights to upscale the VidSet4 sequence. Finally, the model weights of SRGAN being publicly unavailable, we use a third party source (https://github.com/leftthomas/SRGAN) to load the SRGAN model and evaluate its performance on scale factors 2 and 4. Given the super-resolved frames provided by the various models we wish to evaluate, we compute the PSNR, SSIM and PercepDist metric for each model. The results of our computations are shown in table III.

Multiple observations can be made from table III. First, we find that our VSRResNet model, without applying motion compensation on its input, surpasses the state-of-the-art NN-based systems for scale factors of 2 and 3, both in terms of PSNR and SSIM. We argue that the VSRResNet's performance slightly decreases for scale factor of 4 due to the lack of helpful motion details which occurs as a result of the large downscaling factor. In Figures 5 and 6, we provide a comparison of the estimated frames obtained by VSRResNet and the SPMC-VSR [18] for scale factors of 2 and 4. These qualitative results agree with the values shown in Table III, which together show that the VSRResNet network is more successful at super-resolving frames for the scale factor 2, but not as successful for scale factor 4.

Table III shows that while the VSRResNet model outperforms all other SR models in terms of PSNR for scales 2 and 3, the VSRResFeatGAN model is the favored model for producing visually pleasing frames, as determined by the PercepDist measure. In fact, VSRResFeatGAN outperforms VSRResNet, SPMC-VSR [18], VESPCN [27], and Temporal Adaptive Net [28] networks consistently across all scale factors (in addition to surpassing the models trained for still image SR). In addition, its PercepDist measure is much higher than that of the SRGAN model, which also uses a GAN-based training meant to increase the perceptual quality of the resulting images.

To directly assess the qualitative effect of perceptual and adversarial losses, we show in figure 7 a comparison of selected test Myanmar frames estimated by VSRResNet versus those obtained by VSRResFeatGAN for the scale factor 4. Interestingly, while the VSRResNet has a higher PSNR value that VSRResFeatGAN for this scale factor (table III), the frames of VSRResFeatGAN look much sharper. Fine details and textures are recovered in the case of VSRResFeatGAN, whereas these regions are left blurry by VSRResNet. With the significant increase in sharpness, one may note that a subtle amount of distortion is simultaneously introduced in the super-resolved frames of VSRResFeatGAN. This is not an unexpected behavior of GAN models, as these models tend to generate small artifacts, even when strongly regularized with perceptual losses like ours. It may be the case that these artifacts originate from the feature loss's tendency to generate strong grid-like patterns in the frames, which are then further used and distorted by the GAN to provide high-frequency information to the discriminator. The question of whether the subtle artifacts in the frame cancel out the visually pleasing effect of the sharpening is subjective and left to the reader's own assessment. However, even if undesired artifacts are introduced in the outputted frame, the PercepDist still favors the VSRResFeatGAN solution over the traditional MSE-based approaches. We conclude from these qualitative and quantitative results that the use of a combination of perceptual and adversarial losses can have a very significant impact on the resulting quality of the frame.

| | VSRResNet PSNR (dB)/SSIM | VSRResFeatGAN PSNR (dB)/SSIM | SPMC-SR PSNR (dB)/SSIM | VDSR PSNR (dB)/SSIM | SRGAN PSNR (dB)/SSIM | VESPCN PSNR (dB)/SSIM | Temporal Adaptive Net PSNR (dB)/SSIM |
|---|---|---|---|---|---|---|---|
| 2 | **31.87/0.9426** | 30.90/0.9241 | 30.92/0.9235 | 31.61/0.9335 | 29.24/0.9128 | × | × |
| 3 | **27.80/0.8571** | 26.53/0.8148 | 27.49/0.84 | 26.65/0.8091 | × | 27.25/0.8253 | × |
| 4 | 25.51/0.7530 | 24.50/0.7023 | **25.63/0.7709** | 25.05/0.7292 | 23.58/0.7050 | 25.35/0.7309 | 25.53/0.7475 |

| | VSRResNet PercepDist [2] | VSRResFeatGAN PercepDist | SPMC-SR PercepDist | VDSR PercepDist | SRGAN PercepDist | VESPCN PercepDist | Temporal Adaptive Net PercepDist |
|---|---|---|---|---|---|---|---|
| 2 | 0.0407 | **0.0283** | 0.0899 | 0.0541 | 0.0463 | × | × |
| 3 | 0.1209 | **0.0668** | × | 0.1355 | × | 0.1533 | × |
| 4 | 0.1766 | **0.1043** | 0.1908 | 0.1860 | 0.1626 | 0.2022 | 0.1798 |

TABLE III: Comparison with state-of-the-art for VidSet4 dataset on scale factors 2, 3, and 4. The first table uses PSNR and SSIM and the second table uses the Perceptual Distance as defined in [2]. Smaller Perceptual Distance metrics implies better perceptual quality. We were unable to compute the PerceptDist metric for the SPMC-SR model [18] for scale factor 3 as these model weights were not made available by the authors.

Furthermore, not only does the VSRResFeatGAN model outperforms the VSRResNet model in terms of the PerceptDist measure, Table III shows that it also outperforms by a large margin the state-of-the-art VSR models in [18], [28], and [27]. In Figures 8 and 9, we qualitatively compare the results of VSRResFeatGAN with those provided by the SPMC-VSR [18] and the Temporal Adaptive Net in [28]. We focus on these two VSR models as they provide the best PSNR and PercepDist metrics after VSRResFeatGAN in Table III. The particularly sharp quality observed in the VSRResFeatGAN frames is consistent with its superior metrics seen in Table III. More specifically, the 'walk' frame in Figure 8 shows that the solution obtained by VSRResFeatGAN is less blurry than the frame provided by SPMC-SR [18] and the Temporal Adaptive Net [28]. The small distortions introduced from the adversarial loss are most evident in the 'city' frame (Figure 9), in which a strong dot-like pattern is added to the straight lines defining the building architectures. In Figures 10 and 11, we compare the VSRResFeatGAN models with state-of-the-art still image SR models, more particularly the VDSR [29] and the SRGAN [10] networks. The frame in Figure 10 shows the success of our network in sharpening the text printed on the calendar. Finally, the car and leafy trees in Figure 11 are less blurry than the solution proposed by the VDSR [29] and the SRGAN [10] networks.

While the lower PSNR/SSIM values for VSRResFeatGAN may imply lower performance, the qualitative comparison described above, in addition to the figures provided in this paper, clearly suggest that using the PSNR/SSIM metrics may not always accurately assess the performance of a SR model. Instead, our results show that using a learning-based perceptual metric such as the PercepDist introduced here may be more appropriate for comparing the various models in a fairly manner. The significantly low values of the PercepDist metric for the VSRResFeatGAN models imply that it is largely successful at providing SR estimates of visually pleasing quality, and more importantly, is consistent with the observations made from the qualitative results shown in the figures of this paper. Overall, our quantitative and qualitative results show the benefits of using an adversarial approach for training DNNs for VSR along with strong regularization with the use of feature-based losses.

### D. Training Observations

To conclude this section, we describe in more detail the effect of using the Charbonnier loss as a substitute to the $l_2$ loss, and using the weights from VSRResFeatGAN-u2 instead of those of the VSRResNet model for scale factor 4. We show that these training design decisions improve the adversarial training procedure and lead to frames of higher perceptual quality.

#### 1) The effect of the Charbonnier loss during training:
We introduced the use of the Charbonnier loss to measure the distance between an estimated patch and its corresponding ground truth in pixel and feature-space in Section III-C. Other works have chosen to use the $l_2$ loss $\|\hat{x} - G_\theta(Y)\|_2^2$ for

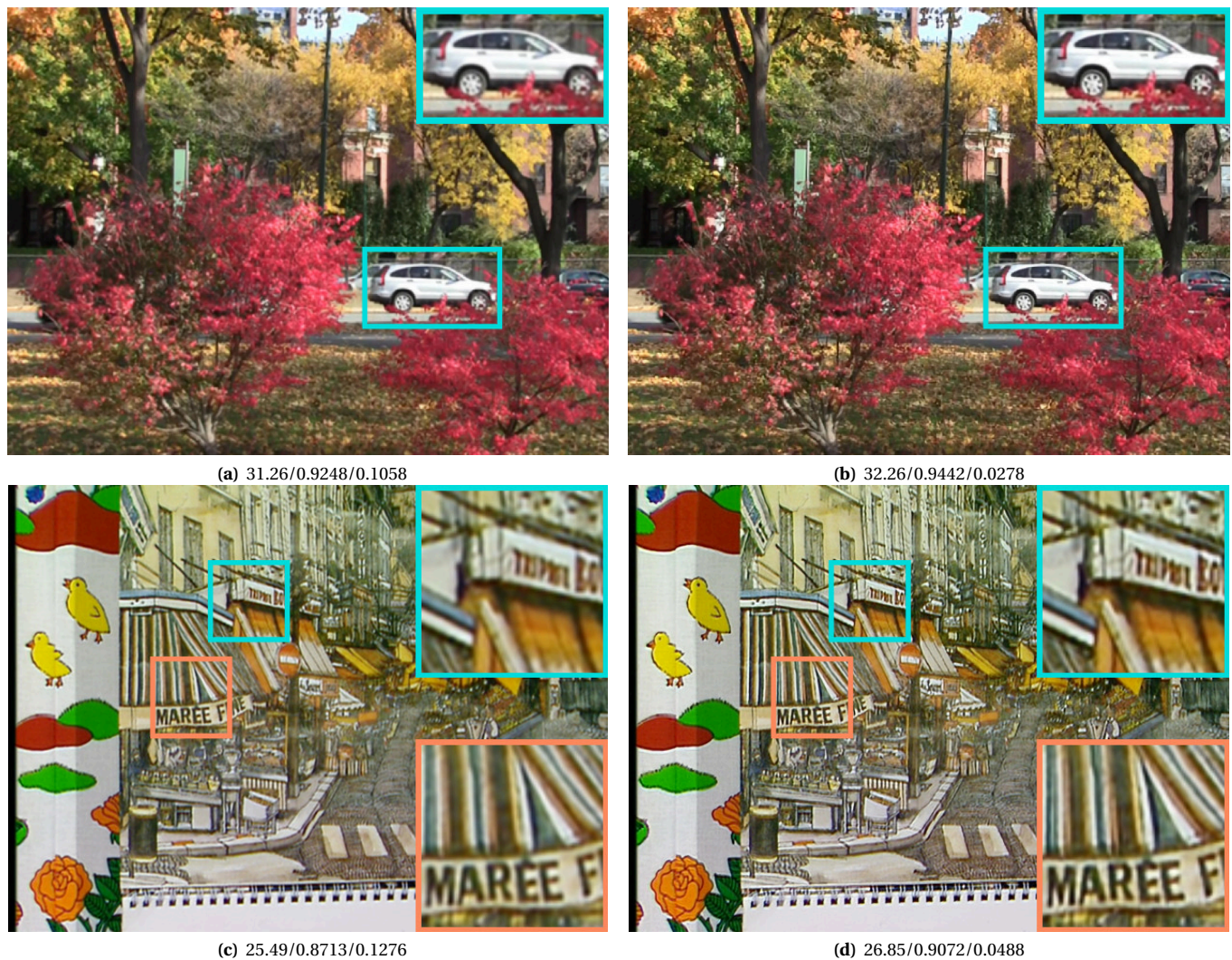

**(a)** 31.26/0.9248/0.1058

**(b)** 32.26/0.9442/0.0278

**(c)** 25.49/0.8713/0.1276

**(d)** 26.85/0.9072/0.0488

**Figure 5:** Qualitative comparison of SPMC [18] (figures (a), (c)) vs. VSRResNet (figures (b), (d)) with zoomed-in regions of VidSet4 for scale factor two. The three numbers in the caption underneath each frame correspond to the PSNR/SSIM/PercepDist metrics computed for that frame.

computing this distance. We compare these two approaches in a controlled setting by replacing the Charbonnier loss term in equation 3 with the $l_2$ distance. We evaluate the effect of each by plotting the loss function of the discriminator during adversarial training, which we show in Figure 12. As demonstrated in Figure 12, the Charbonnier loss provides a more stable alternative to the adversarial training than the $l_2$ loss. With the use of the latter loss function, the discriminator loss converges to values near zero. This indicates that using the $l_2$ loss to regularize the generator results in the network generating patterns which are easily detected by the discriminator. On the other hand, when using the Charbonnier loss, the discriminator loss decreases much more steadily, which implies successful learning between the two adversarial networks. Therefore we conclude that the Charbonnier loss acts as an effective regularizer for controlling the learning dynamics between the generator and the discriminator, providing a more stable alternative to the $l_2$ loss. The reason for this may be due to its particular robustness to small details, as observed in the super-resolution literature (e.g., [30]). This ability to better super-resolve small details than the $l_2$ loss is what makes the discriminator's task slightly more difficult, which in return facilitates the subsequent learning of the generator network.

*2) Transfer Learning from VSRResFeatGAN-u2 to VSRResFeatGAN-u4:* When training VSRResFeatGAN for super-resolving HR frames downsampled by scale factor 4, we found that the quality in the frame estimated by VSRResFeatGAN could greatly improve by initializing the training process with weights obtained from the trained VSRResFeatGAN for the SR task of scale factor 2. This form of transfer learning places the VSRResFeatGAN weights at a good position in parameter space at the beginning of the training process, providing the generator with more leeway to learn accurate SR functions with fewer artifacts. We show the visual effect of setting the initial weights of the VSRResFeatGAN trained for scale factor 2

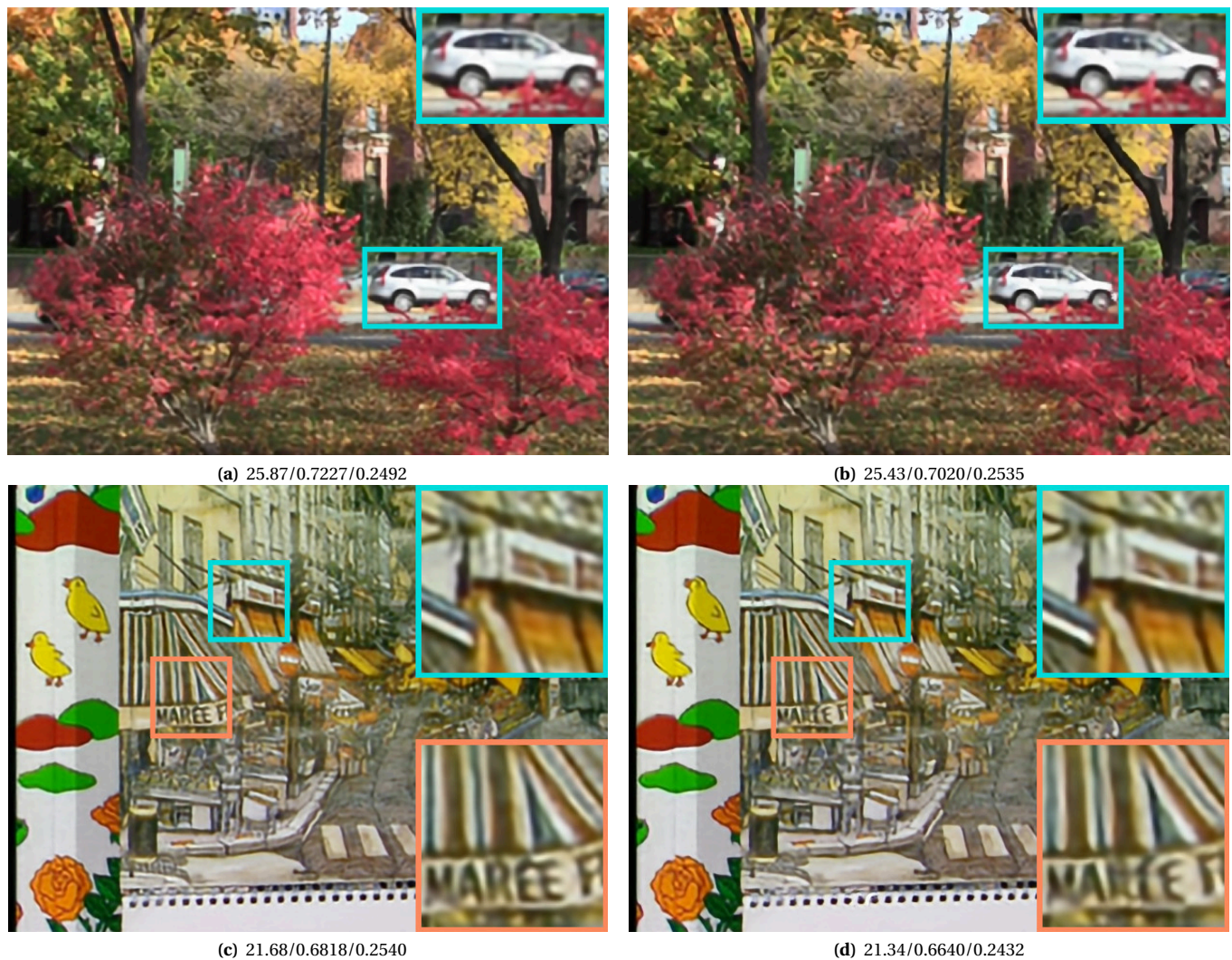

**(a)** 25.87/0.7227/0.2492

**(b)** 25.43/0.7020/0.2535

**(c)** 21.68/0.6818/0.2540

**(d)** 21.34/0.6640/0.2432

**Figure 6:** Qualitative comparison of SPMC [18] (figures (a), (c)) vs. VSRResNet (figures (b), (d)) with zoomed-in regions of VidSet4 for scale factor four.

to that of scale factor 4 in Figure 13. The figure reveals that much of the strong dot-like pattern originating from the adversarial loss is attenuated when using weight transferring from the VSRResFeatGAN training with scale factor 2. Figure 14 shows a comparison of the discriminator and generator loss functions during training. The larger values of the loss function for the discriminator shows that the network has a slightly harder time distinguishing the generated HR patches from the ground truth ones. This implies that the generator generates patches of higher perceptual quality, resulting in the loss function with smaller values, as seen in the right part of Figure 14, which shows the generator's loss function. In conclusion, these experiments reveal that by appropriately trasferring weights from a VSRResFeatGAN trained for a smaller scale factor can greatly help the training of GAN models for more challenging scale factors, which results in frames with fewer high-frequency distortions than when not using weight transferring.

## VI. Conclusion

In this paper, we have shown that training a deep residual neural network with appropriate architectural and loss function choices results in a significant increase in performance, whilst removing the need to perform motion compensation and instead encourage the network to use the motion information for providing better SR estimates. We have applied perceptual losses to video super-resolution by training our deep residual network with GAN losses and Charbonnier distance in feature and pixel spaces. We showed that these losses enabled the network to produce high-resolution frames of significantly higher perceptual quality. In addition to using the PSNR and SSIM metrics for comparing VSRResFeatGAN with current state-of-the-art models, we used the Perceptual Distance metric ([2]) to provide a comparison of the solutions provided by various super-resolution models. We found out that frames of sharp quality that would have been qualified as blurry according to the PSNR metric would in fact achieve a high score with the Perceptual Distance metric.

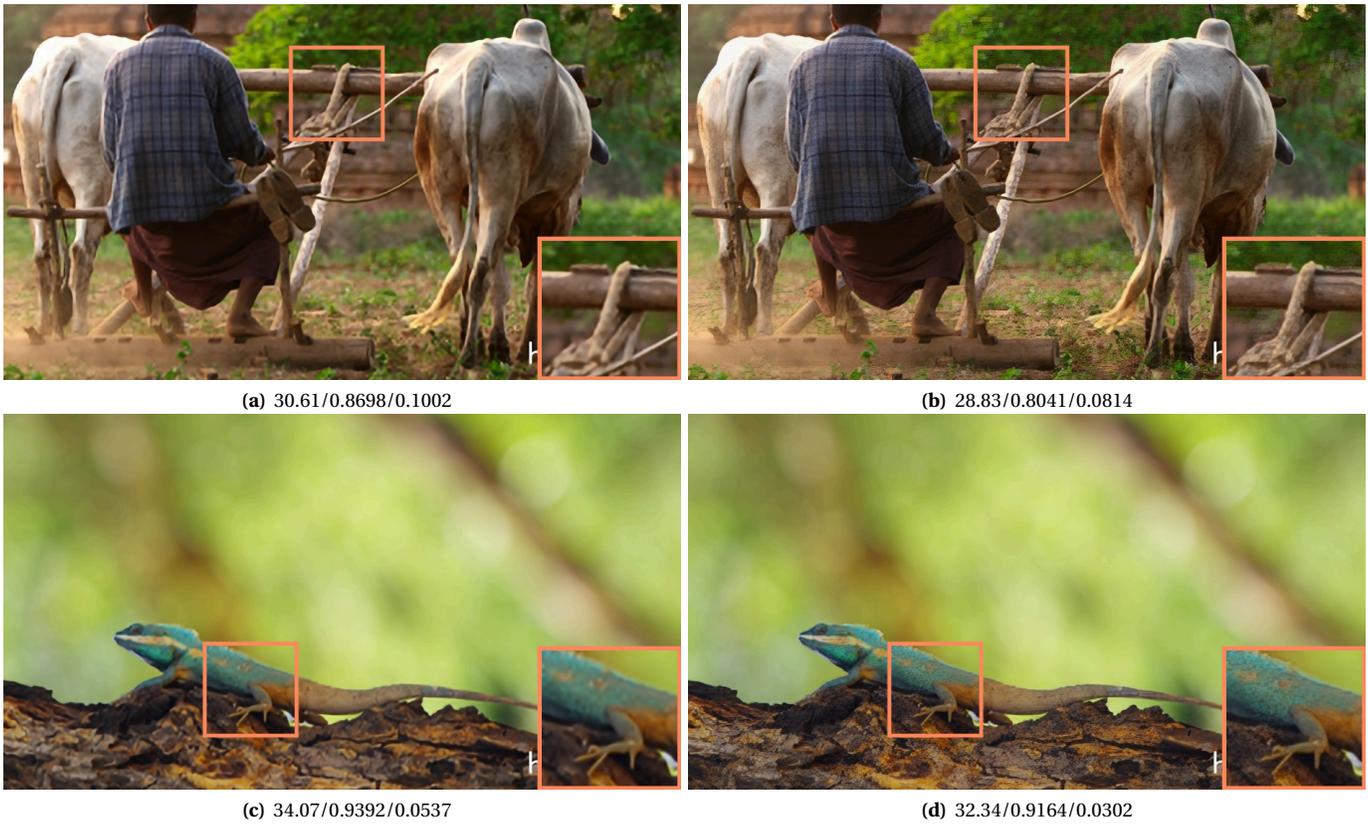

**(a)** 30.61/0.8698/0.1002       **(b)** 28.83/0.8041/0.0814

**(c)** 34.07/0.9392/0.0537       **(d)** 32.34/0.9164/0.0302

**Figure 7:** Qualitative comparison of VSRResNet (figures (a), (c)) vs. VSRResFeatGAN (figures (b), (d)) for scale factor 4 with zoomed-in regions of the Myanmar frames.

While the VSRResNetGAN model is successful at providing sharpened SR estimates to a large degree, it could be further improved by reducing the noise introduced in the estimated frames as a result of the adversarial training. One solution to this would be to constrain the VSRResFeatGAN model to learn SR mappings that are consistent with the mathematical formulation of the VSR problem at hand. This approach may be seen as a way to combine the SR knowledge explicitly used by analytical methods with the powerful ability of GAN-based neural networks, to produce solutions of pleasing visual quality.

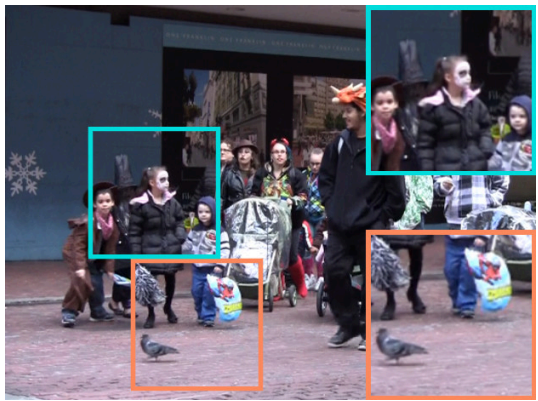 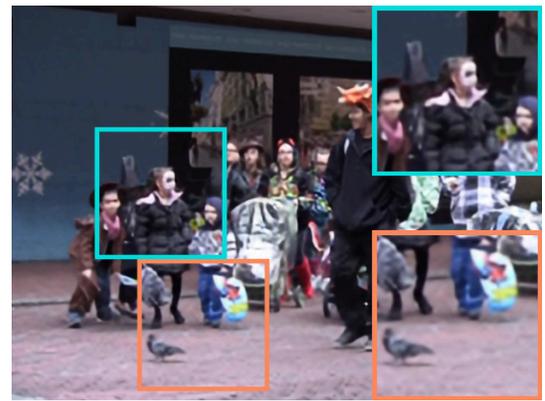

**(a)**          **(b)** 28.83/0.8696/0.0943

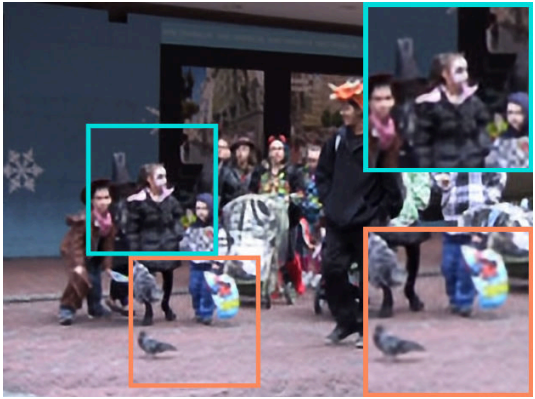 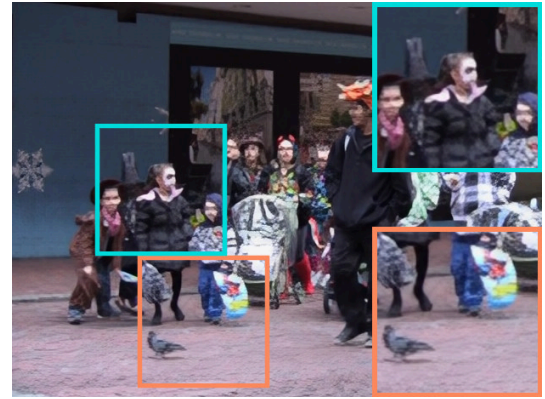

**(c)** 28.37/0.8599/0.0888          **(d)** 27.39/0.8343/0.0457

**Figure 8:** Qualitative comparison of (a) ground truth frame, (b) SPMC-VSR [18] vs. (c) Temporal Adaptive Net [28] and (d) VSRResFeatGAN for scale factor 4, with zoomed in regions. The frame belongs to the 'walk' sequence of the VidSet4 dataset.

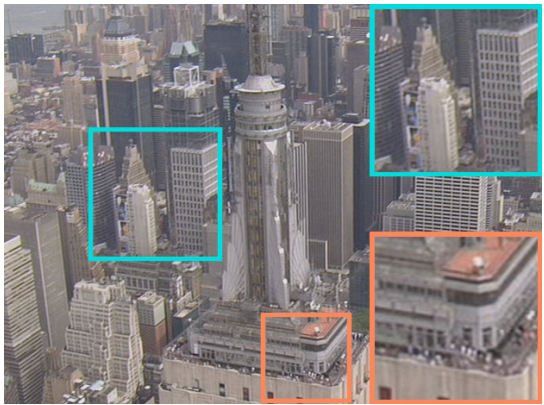 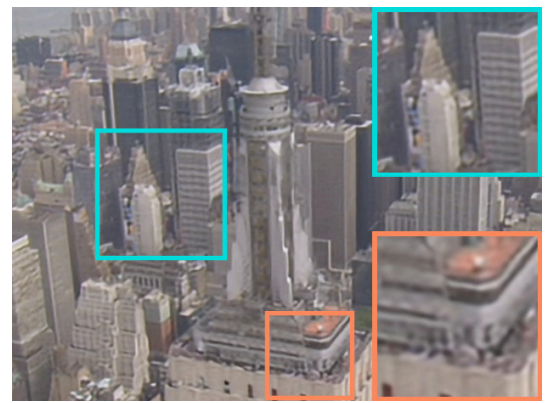

**(a)**          **(b)** 26.56/0.7303/0.1691

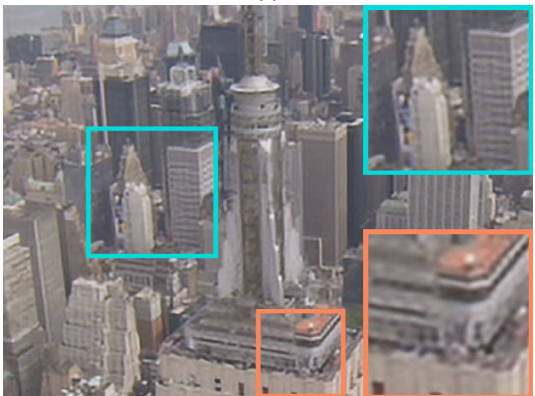 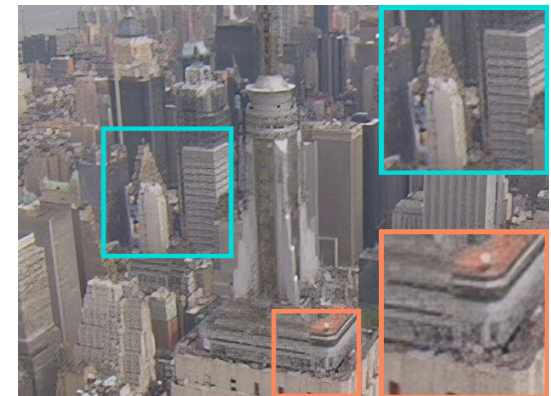

**(c)** 26.32/0.7128/0.1782          **(d)** 25.53/0.6652/0.1305

**Figure 9:** Qualitative comparison of (a) ground truth frame, (b) SPMC-VSR [18] vs. (c) Temporal Adaptive Net [28] and (d) VSRResFeatGAN for scale factor 4, with zoomed in regions. The frame belongs to the 'city' sequence of the VidSet4 dataset.

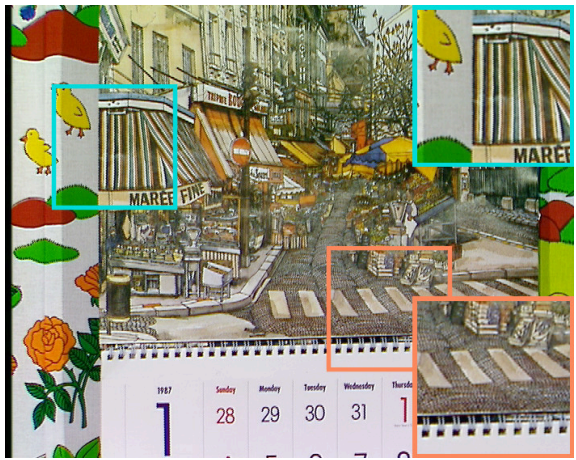 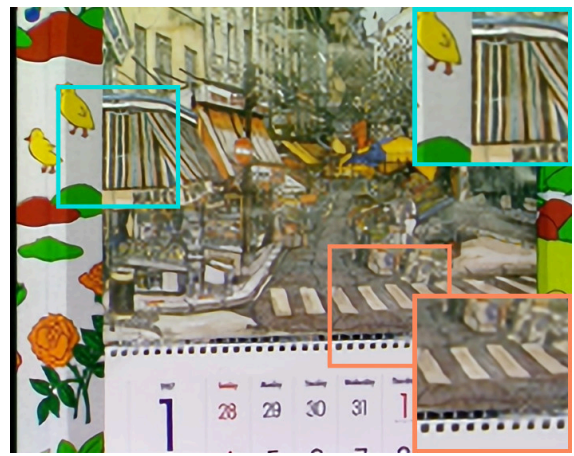

**(a)**  **(b)** 21.33/0.6658/0.2138

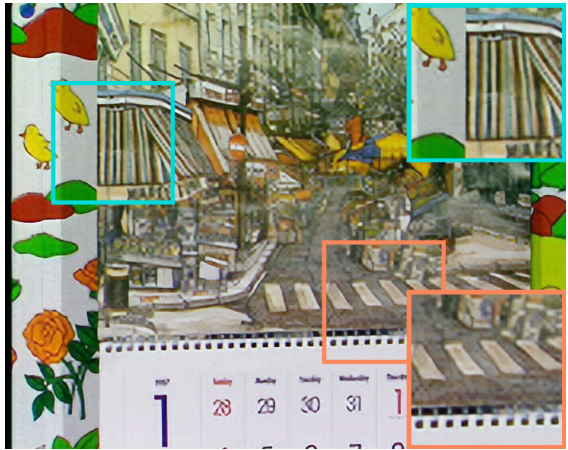 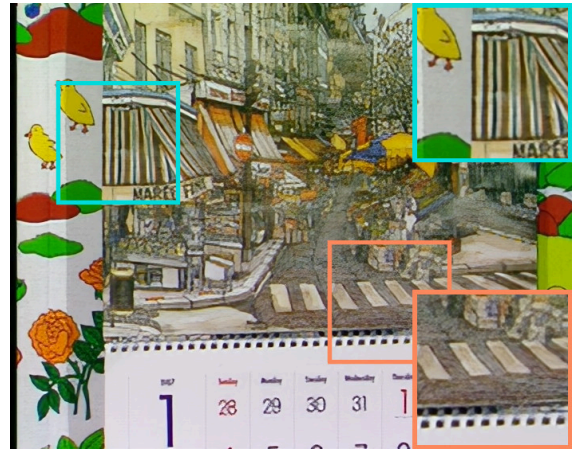

**(c)** 21.37/0.6547/0.1907  **(d)** 21.68/0.7011/0.1272

**Figure 10:** Qualitative comparison of the (a) ground truth frame, (b) VDSR [29], (c) SRGAN [10] vs. (d) VSRResFeatGAN for scale factor 4, with zoomed in regions. The frame belongs to the 'calendar' sequence of the VidSet4 dataset.

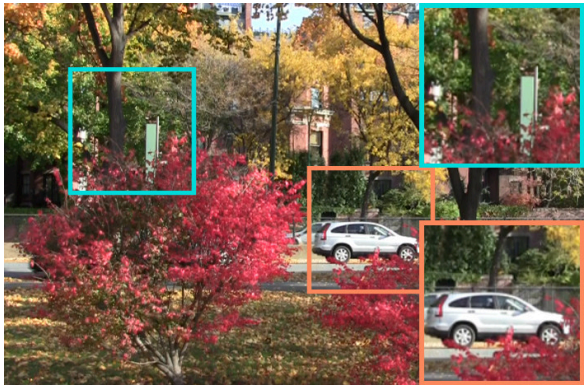 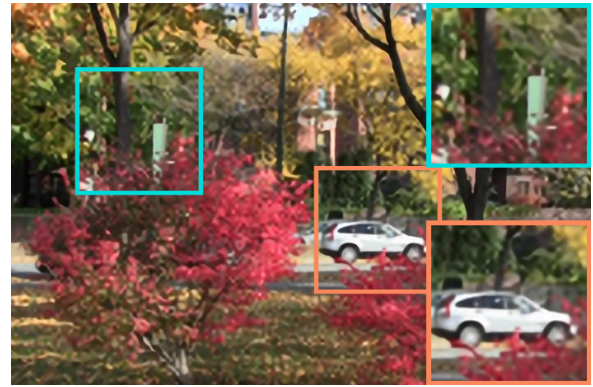

**(a)**  **(b)** 24.51/0.6469/0.2611

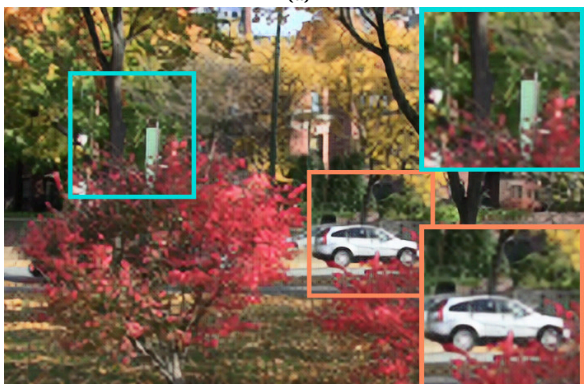 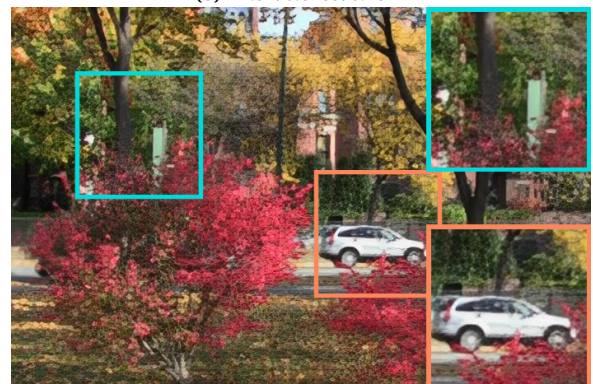

**(c)** 24.34/0.6375/0.2281  **(d)** 23.09/0.5694/0.1259

**Figure 11:** Qualitative comparison of the (a) ground truth frame, (b) VDSR [29], (c) SRGAN [10] vs. (d) VSRResFeatGAN for scale factor 4, with zoomed in regions. The frame belongs to the 'foliage' sequence of the VidSet4 dataset.

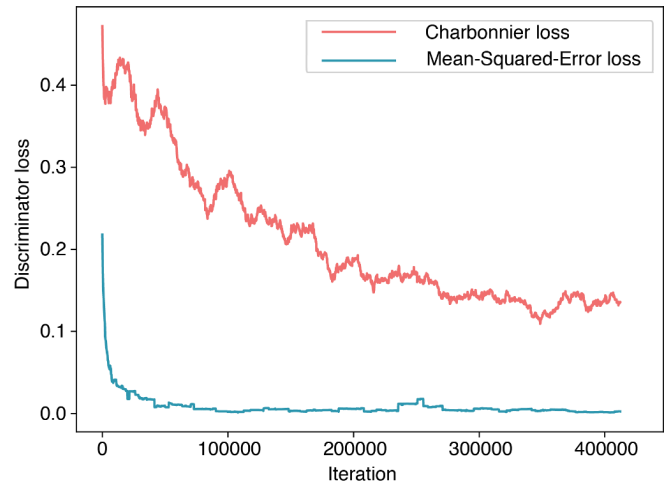

**Figure 12:** The effect of using the Charbonnier loss instead of the standard MSE as regularization during adversarial training leads to a more stable training procedure. When using MSE, the discriminator quickly learns to distinguish super-resolved patches from ground truth ones, which results in limited learning for the generator.

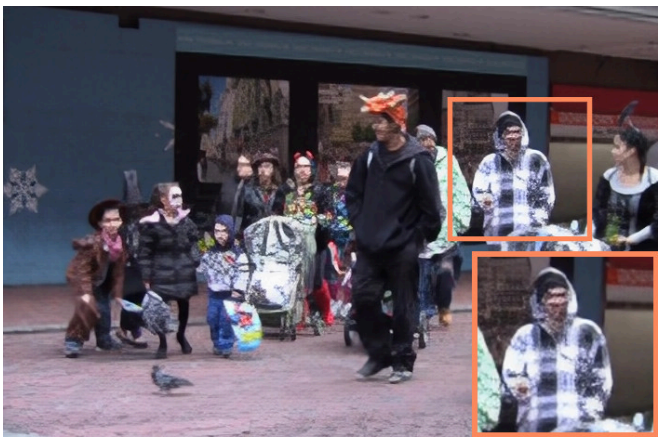
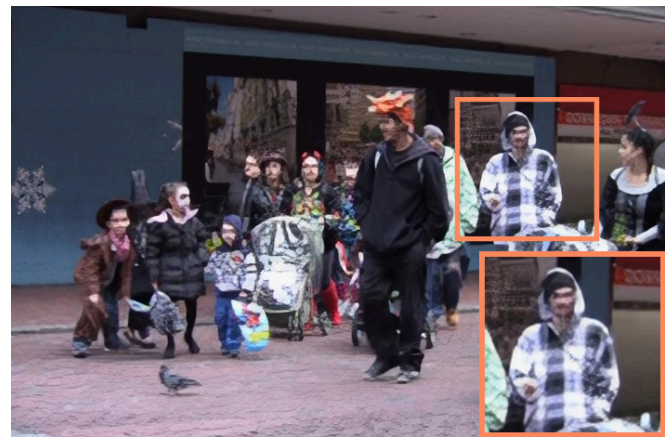

**(a)** 26.62/0.8167/0.0539  **(b)** 27.39/0.8354/0.0457

**Figure 13:** A comparison of initial VSRResFeatGAN's weights for scale factor 4 (figure (a)) with those of the converged VSRResFeatGAN for scale factor 2 (figure (b)), with zoomed in regions. The dot-like artifacts originating from the GAN training are greatly attenuated when using weight transferring from scale factor 2.

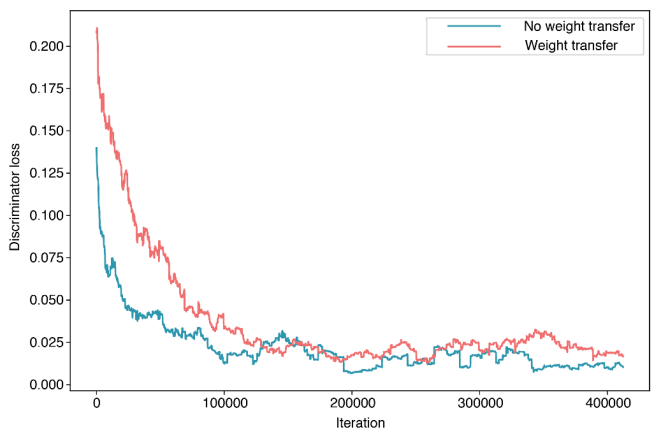
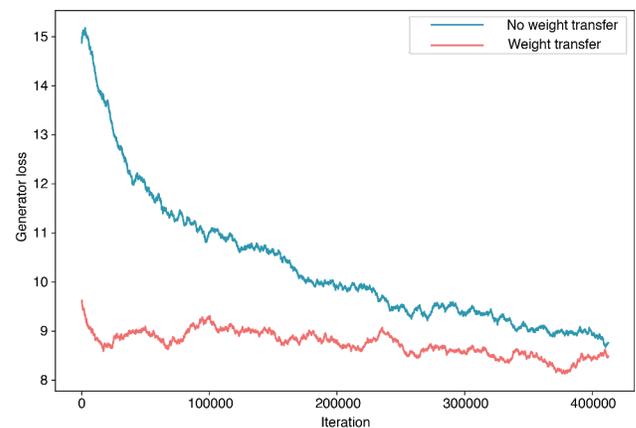

**Figure 14:** The effect on the loss functions of the discriminator (left) and generator (right) of using the weights from VSRResFeatGAN pre-trained on scale factor 2, vs. starting with the VSRResNet weights pre-trained on scale factor 4.